\documentclass[conference]{IEEEtran}
\IEEEoverridecommandlockouts
\usepackage{cite}
\usepackage{amsmath,amssymb,amsfonts}
\usepackage{algorithmic}
\usepackage{graphicx}
\usepackage{epsfig}
\usepackage{epstopdf}
\usepackage{textcomp}
\usepackage{xcolor}
\usepackage{float}
\def\BibTeX{{\rm B\kern-.05em{\sc i\kern-.025em b}\kern-.08em
    T\kern-.1667em\lower.7ex\hbox{E}\kern-.125emX}}
\begin{document}

\title{Knowledge-Aided Semantic Communication Leveraging Probabilistic Graphical Modeling\\
% \thanks{This work is partly supported by NSFC under grant No. 62293481, No. 62201505, partly by the SUTD-ZJU IDEA Grant (SUTD-ZJU (VP) 202102)}
}
\author{\IEEEauthorblockN{Haowen Wan$^{1,2}$, Qianqian Yang$^{2}$, Jiancheng Tang$^{2}$, Zhiguo Shi$^{1,2,3}$}
\IEEEauthorblockA{$^1$ Polytechnic Institute, Zhejiang University, Hangzhou, China}
\IEEEauthorblockA{$^2$ College of Information Science and Electronic Engineering, Zhejiang University, Hangzhou, China}
\IEEEauthorblockA{$^3$ Jinhua Institute of Zhejiang University, Jinhua, China}
\IEEEauthorblockA{Email:\{haowenwan20, qianqianyang20, jianchengtang, shizg\}@zju.edu.cn}
}

\maketitle
\thispagestyle{empty}
\pagestyle{empty}

\begin{abstract}
In this paper, we propose a semantic communication approach based on probabilistic graphical model (PGM). 
The proposed approach involves constructing a PGM from a training dataset, which is then shared as common knowledge between the transmitter and receiver. We evaluate the importance of various semantic features and present a PGM-based compression algorithm designed to eliminate predictable portions of semantic information. Furthermore, we introduce a technique to reconstruct the discarded semantic information at the receiver end, generating approximate results based on the PGM. Simulation results indicate a significant improvement in transmission efficiency over existing methods, while maintaining the quality of the transmitted images.
\end{abstract}

\begin{IEEEkeywords}
Semantic communication, Image transmission, Probabilistic graphical model  
\end{IEEEkeywords}

\section{Introduction}
Data-intensive applications like ultra-high-definition (UHD) video transmission, autonomous driving, mixed reality (MR), and the metaverse are driving a surge in wireless data traffic, posing challenges for existing wireless systems \cite{ref1}. Semantic communication (SemCom) stands out as a promising technology that greatly enhances transmission efficiency by transmitting compact semantic information while discarding semantically insignificant content. In contrast to conventional communication systems, which prioritize transmitting symbols while ignoring their underlying meaning, SemCom approaches focus on transmitting and retrieving task-relevant information rather than aiming for precise bit recovery. SemCom has demonstrated its superiority in terms of transmission efficiency, particularly in harsh channel conditions.\cite{ref2}.

Driven by the advantages offered by SemCom, existing work harnesses recent advancements in deep learning techniques to achieve remarkable performance in transmitting various signals such as text\cite{ref3}, speech\cite{ref4}, image\cite{ref5, ref6, ref7}, and video\cite{ref8} signals.  In these DL-based semantic communication systems, we can interpret the neural network (NN) weights as knowledge acquired from the training dataset, which is utilized for extracting and recovering semantic information. Typically, the semantic information is perceived as the intermediate latent features output by specific neural network layers. In this way, the semantic information varies in both formats and values when a different NN model is used, lacking a coherent and easily interpretable representation, thereby weakening the generalization ability of SemCom systems across various transmission tasks.

To establish a comprehensive SemCom framework, defining a unified representation of semantic information is crucial. Existing approaches include propositional logic, knowledge graphs, and Probabilistic Graphical Models (PGMs), among which PGMs stand out as particularly promising.  PGMs are a widely used graphical method for depicting statistical relationships among variables. A PGM consists of nodes, edges, and parameters: nodes represent variables or real-world objects, edges denote relationships between nodes, and parameters capture probabilistic factors. PGMs are extensively employed in the semantic analysis and comprehension of various data types, including text \cite{ref9}, images, and video \cite{ref10}, demonstrating their generality in characterizing semantic information. Motivated by this, several studies have integrated PGMs into SemCom system design. For instance, \cite{ref11} proposed a SemCom system for text transmission utilizing PGMs to represent shared knowledge about text data between transmitters and receivers. This approach enabled text compression at the transmitter and subsequent recovery at the receiver, reducing signal redundancy. However, this method is limited to text transmission, highlighting the need for further exploration of PGM-based SemCom approaches for other transmission tasks.

In this paper, we propose a SemCom approach based on PGM, where a PGM is trained on data and employed as shared semantic knowledge between the transmitter and receiver to minimize transmission redundancy. Specifically, we introduce a PGM-based compression algorithm for semantic information by eliminating easily predictable features through PGM inference, while considering the importance of these features. The receiver then restores the discarded semantic information through inference with the shared PGM using the received semantic data. To evaluate the effectiveness of the proposed framework, we develop a PGM-based SemCom system for facial image transmission. Specifically, we utilize Semantic StyleGAN \cite{ref12} to extract disentangled semantic features and train a PGM to characterize the statistical correlations between them. Simulation results demonstrate a significant improvement in transmission efficiency compared to existing methods, while preserving the quality of transmitted images.

\begin{figure*}
    \includegraphics[width=6.5in]{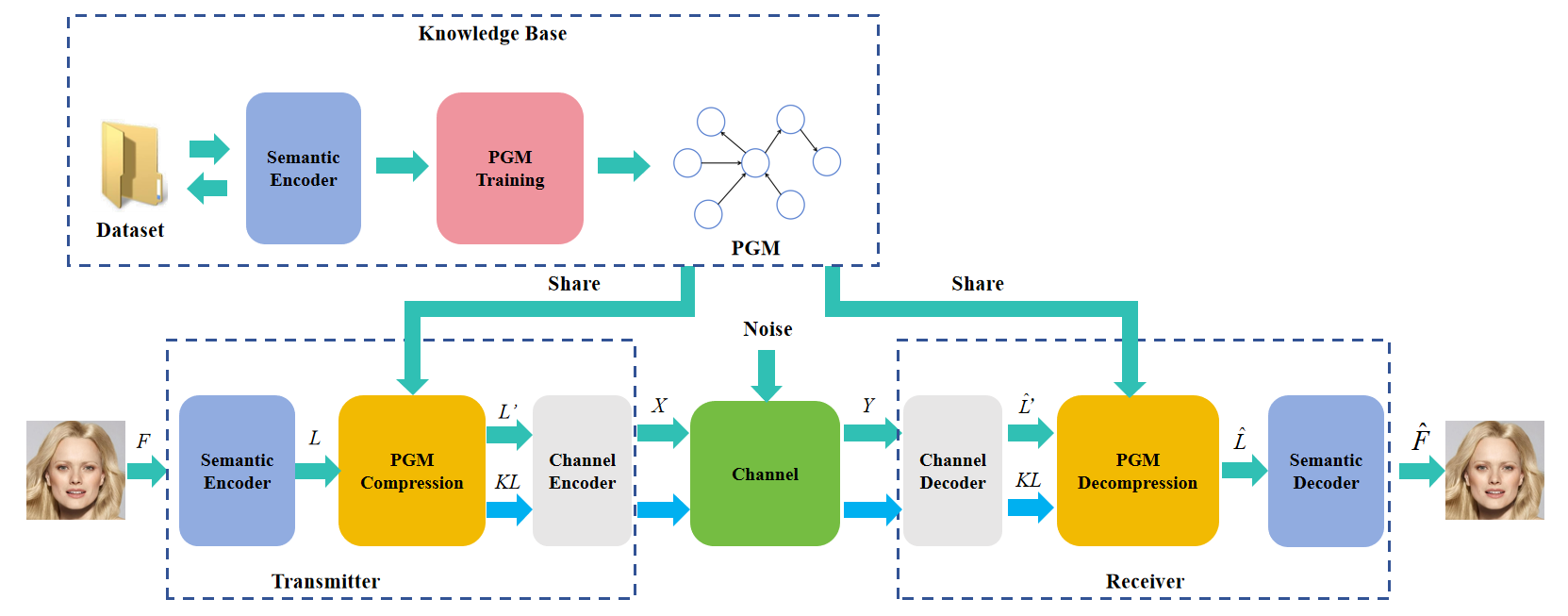}
    \caption{The overall structure of the proposed semantic communication system for image transmission}
    \label{fig:1}
\end{figure*}

% The construction of this paper can be summarized as follows: (i) We introduce a comprehensive SemCom framework based on PGM, utilizing PGM as a shared knowledge base to accurately represent semantic information. (ii) We propose a learning method for constructing PGMs from training datasets, which results in a complete PGM that can be utilized for inference and prediction. (iii) We propose a compression algorithm that utilizes the learned PGM to further reduce communication overhead while considering the importance of semantic features. (iv) To validate the effectiveness of our framework, we implement the proposed PGM based SemCom for facial image transmission, which employs Semantic StyleGan as the semantic encoder to extract disentangled semantic features.

\section{SYSTEM MODEL}
In this section, we present the system model of the proposed SemCom approach. We also introduce the definition of compression ratio and metrics to evaluate the performance of the proposed model for image transmission.

\subsection{Knowledge Base}
As shown in Fig. 1, we learn a PGM from a given dataset that shares the same distribution as the data to be transmitted later. The original data in the training set are first disentangled into latent representations using a semantic encoder. We then use a Bayesian network (BN), a directed acyclic graph (DAG) composed of nodes and edges, to represent directional causality and conditional probability distributions (CPDs) among the semantic features. The nodes in the BN correspond to semantic features extracted by the semantic encoder, and the directional edges are determined by a structural learning algorithm. The CPDs in the BN are estimated using a parameter learning algorithm. This BN, along with other training results, forms the knowledge base of the proposed model, which is shared by the transmitter and receiver for inference.

\subsection{Transmitter}
The transmitter in the considered SemCom system consists of a semantic encoder, a PGM compression module, and a channel encoder. First, the semantic encoder extracts a compact latent semantic representation, denoted as $\textbf{\textit{L}}$, from the original data $\textbf{\textit{F}}$. The PGM compression module then uses the PGM's inference results to filter out the part of the latent code that can be easily inferred, resulting in a reduced latent representation, denoted as $\textbf{\textit{L}}^{\prime}$. The channel encoder maps $\textbf{\textit{L}}^{\prime}$ into a symbol sequence $\textbf{\textit{X}}$ for transmission over the physical channel. It is crucial for the receiver to know the exact indices of the preserved features, denoted as $\textbf{\textit{KL}}$, to reconstruct the original image. Therefore, $\textbf{\textit{KL}}$ is transmitted losslessly using digital communication techniques.

\subsection{Receiver}
The received signal at the receiver is given by 
\begin{equation}
    \textbf{\textit{Y}} = \textbf{\textit{X}} + \textbf{\textit{N}} 
\end{equation}
where $\textbf{\textit{N}} \sim \mathcal{N}\left(0, \sigma^2\right)$ denotes the Gaussian noise of the channel. The receiver consists of a channel decoder, a PGM decompression module, and a semantic decoder. The received signal $\textbf{\textit{Y}}$ is first mapped back into the latent semantic representation $\hat{\textbf{\textit{L}}^{\prime}}$ by the channel decoder. Then, the PGM decompression module uses $\hat{\textbf{\textit{L}}^{\prime}}$ and their corresponding indices $\textbf{\textit{KL}}$ to recover the discarded latent features through PGM inference. Eventually, the semantic decoder combines $\hat{\textbf{\textit{L}}^{\prime}}$ and the inferred latent features to reconstruct the image $\hat{\textbf{\textit{F}}}$.

\subsection{Metrics}
We evaluate our system's performance for an image transmission task in terms of image reconstruction quality and bandwidth compression ratio. To assess image reconstruction quality, we use two widely adopted metrics: Peak Signal-to-Noise Ratio (PSNR) and Learned Perceptual Image Patch Similarity (LPIPS) \cite{ref13}. PSNR measures the pixel-level difference between two images, denoted by
\begin{equation}
    PSNR=10 \times \log_{10}{\frac{Max(F,\hat{F})^2}{MSE(F,\hat{F})}} 
\end{equation}
where $Max(F,\hat{F})^2$ represents the maximum difference between pixel values in the images. LPIPS measures the distance in the feature space derived by the pre-trained AlexNet from a human perceptual perspective, denoted by
\begin{equation}
    LPIPS(F, \hat{F})=\sum\limits_{l}\frac{1}{H_lW_l}\sum\limits_{i,j} \left\Vert c_l\odot \left(f_{ij}^l-\hat{f_{ij}^l}\right)\right\Vert_2^2 
\end{equation}
where $f^l_{i,j}$ and $\hat{f^l}_{i,j}$ denote the $\left(i,j\right)$th element in normalized latent feature maps output by layer $l$ of the AlexNet, respectively. $H_l$ and $W_l$ denote the height and width of the feature map, respectively. Notation $\odot$ denotes the scale operation and $c_l$ represents the pretrained weights for intermediate latent output by layer $l$.

We employ bandwidth compression ratio to evaluate the transmission efficiency. We define the image dimension $n$ as the source bandwidth, and the channel dimension $k$ as the channel bandwidth. We denote the ratio $k\/n$ as the bandwidth compression ratio. 

\section{Proposed Method}
In this section, we introduce a PGM-based SemCom approach for facial image transmission. We begin by explaining the inversion method of Semantic StyleGAN, followed by the introduction of PGM training. Next, we present the compression and decompression algorithms for the decoupled latent code. Finally, we outline the training procedures for the channel encoder and decoder.

\subsection{Semantic Encoder and Decoder}
To extract disentangled semantic information from input image $x$, we leverage the inversion method\cite{ref14} of Semantic StyleGAN generative function, denoted by $G\left(\cdot\right)$, for semantic encoding and decoding. The semantic encoder aims to find the semantic information of the input image, which can be used to generate an image most similar to the original using the generative function $G\left(\cdot\right)$. This process can be formulated as the following optimization problem:
\begin{equation}
    L = \mathop{\arg\min}\limits_{y} \lambda_1\Vert G\left(y\right)-F \Vert_2^2 + \lambda_2 LPIPS\left( G\left(y\right),F \right) 
\end{equation}
where $L$ represents the extracted disentangled latent information of the input image  $F$, $\lambda_1$ and $\lambda_2$ are hyperparameters used to balance the l2-norm loss and LPIPS loss. This optimization problem can be effectively solved using gradient descent methods. 

The semantic decoder utilizes the generative function of Semantic StyleGAN to decode the recovered latent representation $\hat{L}$. The generation process can be expressed as
\begin{equation}
    \hat{F} = G\left(\hat{L}\right)
\end{equation}
where $\hat{F}$ is the reconstructed image.

\begin{figure}
    \includegraphics[width=3in]{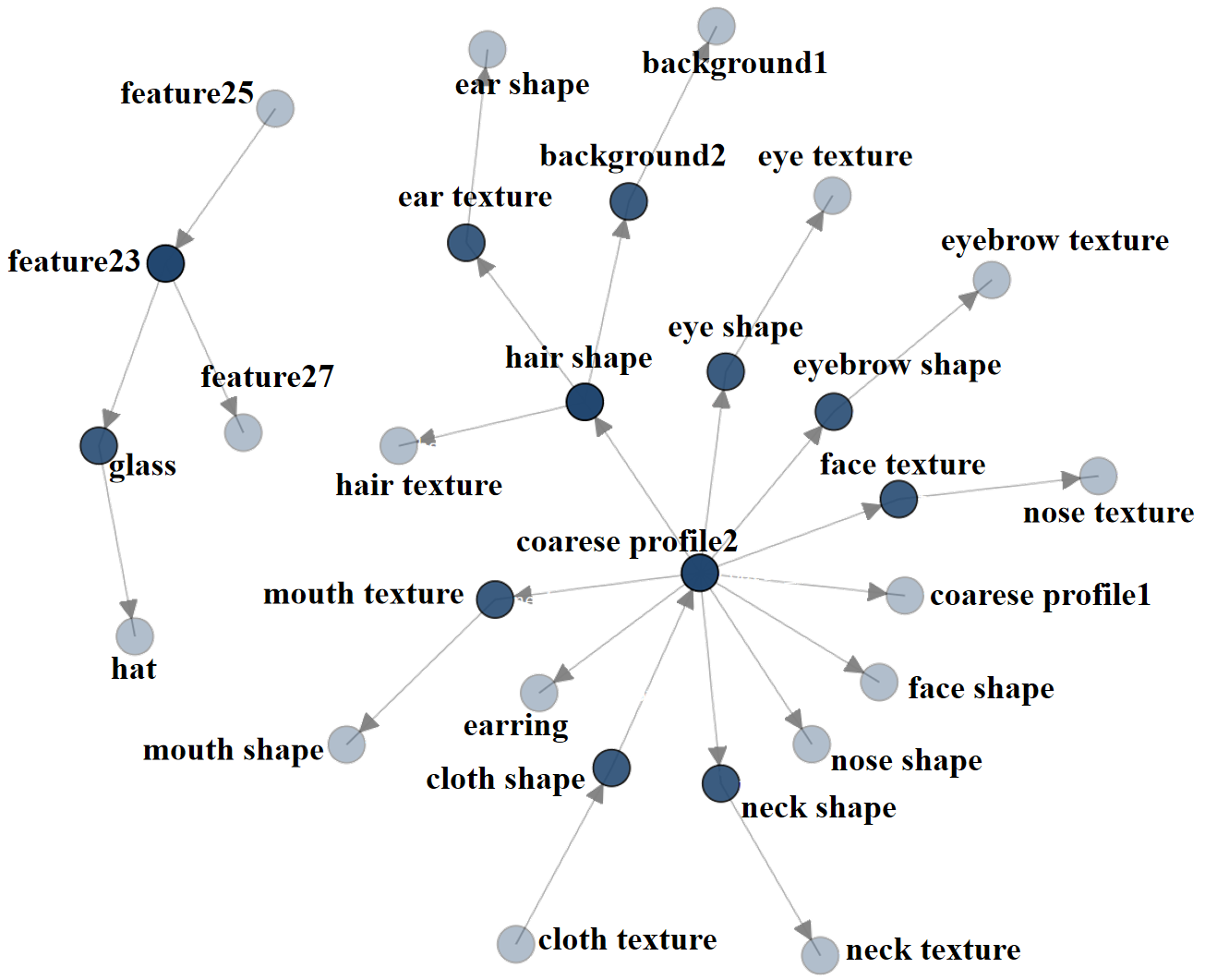}
    \caption{Bayesian network trained on CelebA-HQ dataset}
    \label{fig:2}
\end{figure}

\subsection{Construction of Probabilistic Graphical Model}
In this paper, we utilize the first 28,000 images from the CelebA-HQ dataset \cite{ref15} for training and preprocessing the images in the dataset before constructing the Bayesian network (BN). Given the large scale of the dataset, analysing each latent code one by one not only causes a large computational cost, but also is not conducive to finding the probabilistic relations between semantic features. To address the problem, we consider using the Mini-Batch Kmeans algorithm to compress the domain of each semantic feature.

After the above preparation, we utilize bnlearn\cite{ref16} to complete the parameter learning and structure learning of BN. We choose score-based structure learning and utilize the HillClimbSearch algorithm for constructing BN. As for parameters, we employ Maximum Likelihood Estimation (MLE) for parameter estimation based on training data. During the search, we adopt Bayesian Information Criterion (BIC) for selecting the best model from a finite set of model. Given the dataset $D$, BIC scoring function of Bayesian network $B= \left \langle G,\Theta \right \rangle$ can be written as:
\begin{equation}
    BIC\left(B|D\right)= -\frac{\log N}{2}\left|B\right| + LL\left(B|D\right)
\end{equation}
where $\left|B\right|$ is the number of parameters in BN,  $LL\left(B|D\right)$ is the the logarithmic likelihood of BN. As depicted in Fig.2, we ultimately obtain a Bayesian network trained on the CelebA-HQ dataset for prediction and inference.

% The training of the Bayesian network comprises structure learning and parameter learning. We utilize bnlearn \cite{ref16}, which analyzes statistical data and aids in constructing the Bayesian network. For structure learning, we employ a heuristic search approach called HillClimbSearch. The results of structure learning are evaluated using the Bayesian Information Criterion (BIC). Subsequently, we estimate the parameters of the Bayesian network based on the training data, employing Maximum Likelihood Estimation (MLE) as our parameter learning algorithm. As depicted in Fig.2, we ultimately obtain a Bayesian network trained on the CelebA-HQ dataset for prediction and inference. The PGM is also shared by the transmitter and receiver as side information.

\begin{figure*}
    \centering
    \begin{minipage}[t]{0.43\textwidth}
        \centering
        \includegraphics[width=3in]{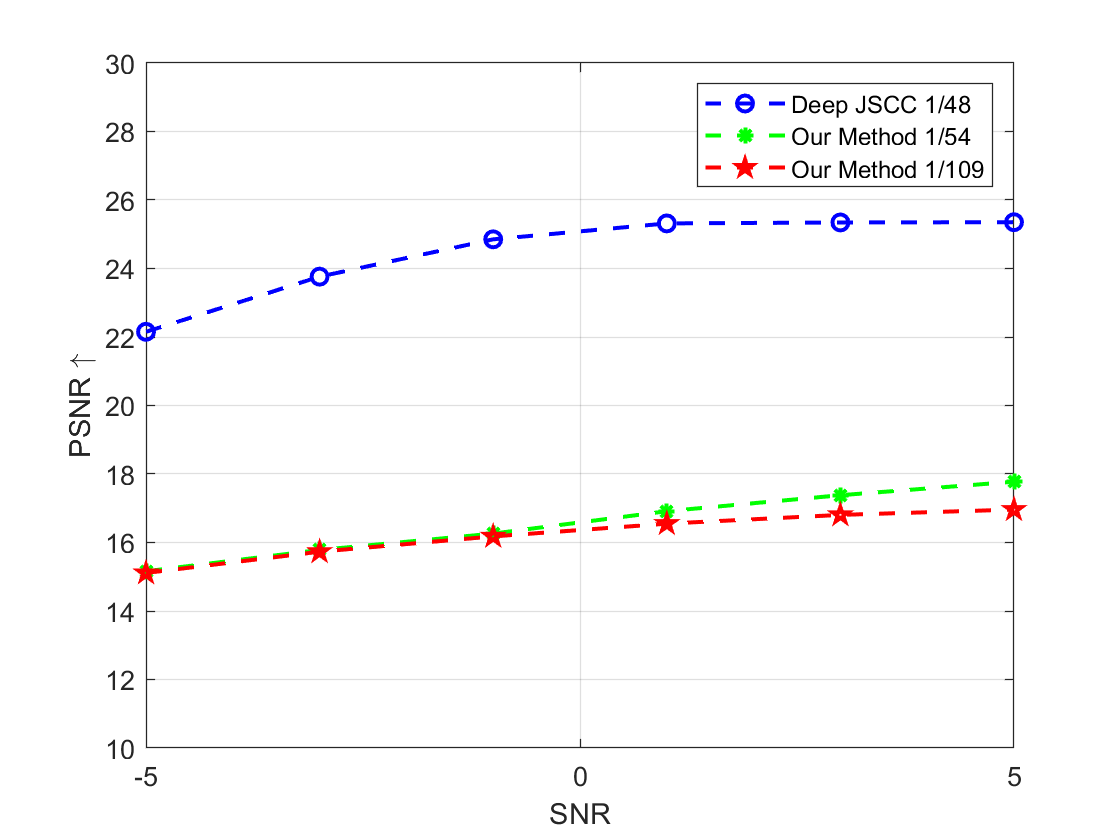}
        \caption{PSNR versus SNR for different approaches. The compression ratios are listed in the legend.}
        \label{fig:3}
    \end{minipage}
    \hspace{0.65in}
    \begin{minipage}[t]{0.43\textwidth}
        \centering
        \includegraphics[width=3in]{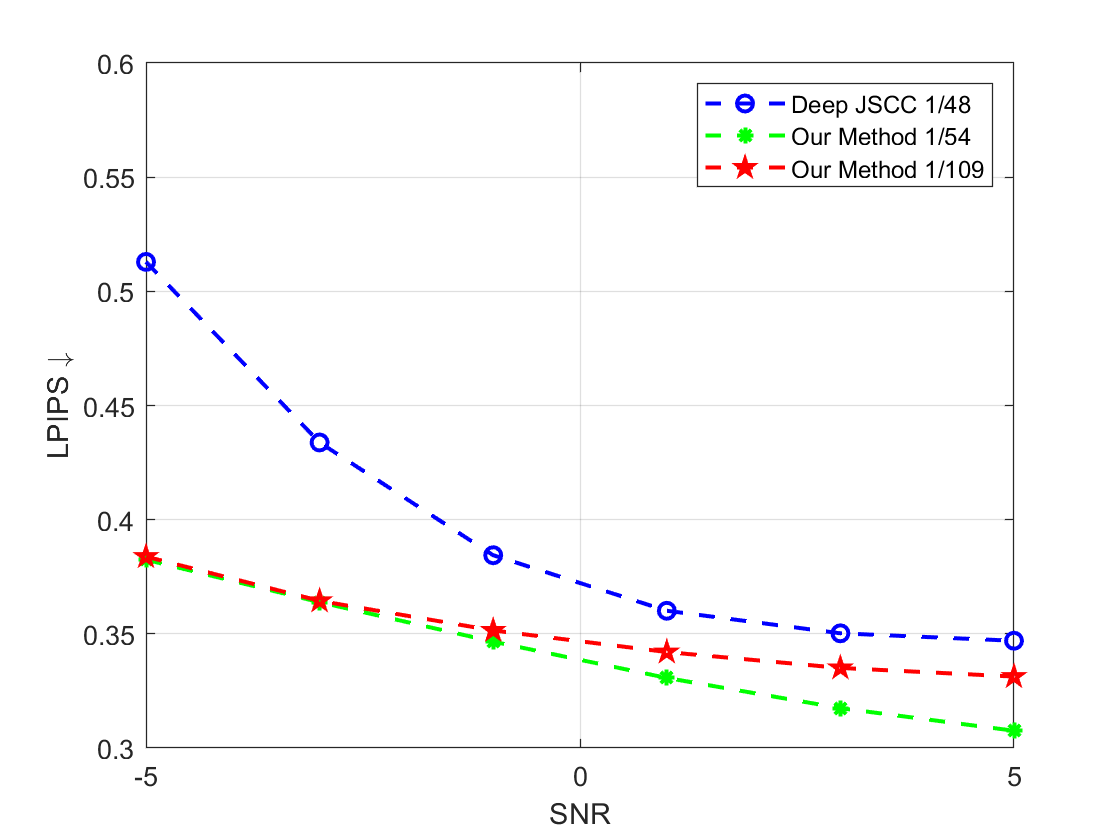}
        \caption{LPIPS versus SNR for different approaches. The compression ratios are listed in the legend.}
        \label{fig:4}
    \end{minipage}
\end{figure*}

\begin{figure*}
    \includegraphics[width=7in]{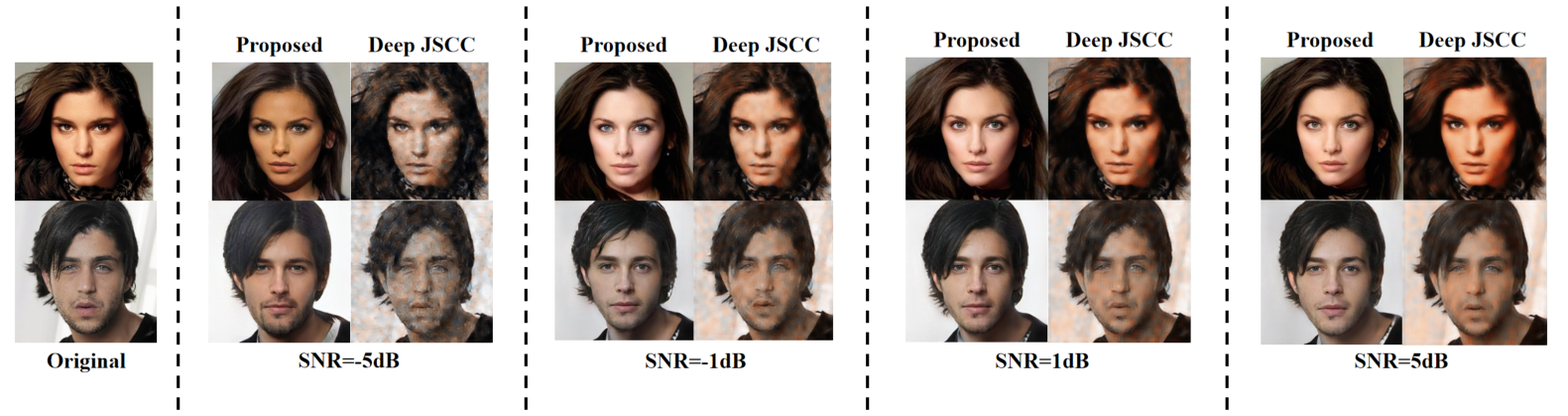}
    \caption{Example of reconstructed images produced by the Deep JSCC and proposed method for image transmission for AWGN channel, and the compression ratio of Deep JSCC and proposed method is 1/48 and 1/109 respectively. From left to right, the columns correspond to SNR values of -5dB, -1dB, 1dB, 5dB.}
    \label{fig:5}
\end{figure*}

\subsection{Compression and Decompression Algorithm}
At the transmitter side, we employ the inversion method of Semantic StyleGAN for semantic information extraction, which also serves as compression for images. Subsequently, we utilize a PGM-based compression algorithm for further compression. Initially, we simulate the inference error of the PGM by replacing it with a random clustering center and measure the LPIPS and MSE to assess the importance of each semantic feature. Based on these results and the structure of the PGM, we retain 14 features as side information for inference and assign weights to the remaining negligible features.

With this preparation, we utilize the PGM to compute the correct inference probability of each negligible node based on other nodes, which can be expressed as:
\begin{equation}
    P_1 = \left [ p_1, p_2, \cdots, p_i, \cdots, p_{14} \right] 
\end{equation}
where $P_1$ is the probabilistic vector of the first round, $p_i$ is the correct inference probability of $i_{th}$ node. After multiplying the probability of each node by its weight, we select the highest weighted probability, denoted by:
\begin{equation}
    p_{max1} = Max\left(W_1 \cdot P_1\right)
\end{equation}
where $W_1$ is the weight vector of the first round. We then find the node corresponding to $p_{max1}$ and discard its dimension in the latent code. This process is repeated several times until we obtain the compressed latent code and the list of retained dimensions.

The compression algorithm's concept is applied to the decompression algorithm, enabling the deduction of missing dimensions by the PGM. At the receiver side, upon receiving the compressed latent code, the system infers the most likely corresponding state of each ignored node based on the PGM. Subsequently, it selects the highest probability $p_{max1}^{\prime}$ in the probabilistic vector, which can be written as:
\begin{equation}
    p_{max1}^{\prime} = Max\left(P_1^{\prime}\right) = Max \left[p_{1}^{\prime}, p_{2}^{\prime}, \cdots, p_j^{\prime}, \cdots, p_{14}^{\prime} \right]
\end{equation}
where $P_1^{\prime}$ is the probabilistic vector of the first round, $p_j^{\prime}$ is the probability of the most likely corresponding state of $j_{th}$ node. The node and its state corresponding to $p_1^{\prime}$ is assumed as known information, and its dimension in latent code is then replaced by the clustering center of its approximate state. This process is repeated until the omitted dimensions in the latent code are recovered, resulting in the decompressed latent code.

\begin{figure*}
    \centering
    \begin{minipage}[t]{0.43\textwidth}
        \centering
        \includegraphics[width=3in]{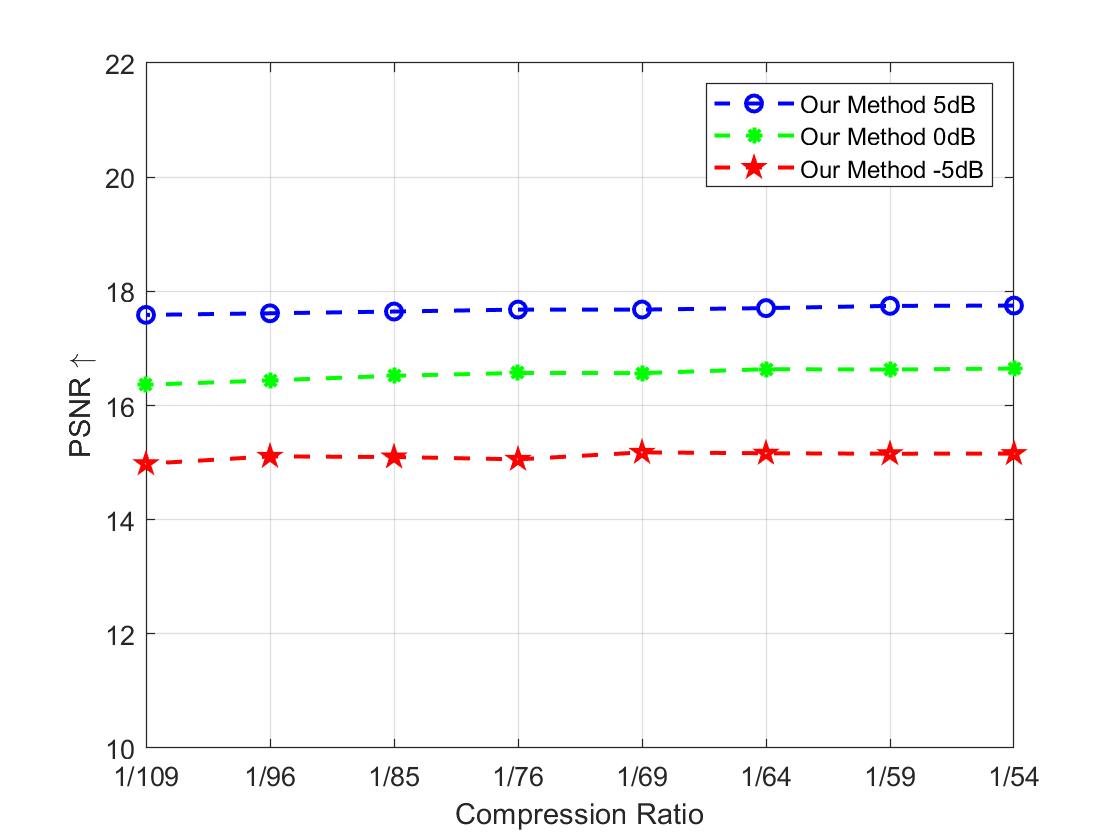}
        \caption{PSNR versus compression ratio, SNR=-5dB, 0dB, 5dB.}
        \label{fig:6}
    \end{minipage}
    \hspace{0.65in}
    \begin{minipage}[t]{0.43\textwidth}
        \centering
        \includegraphics[width=3in]{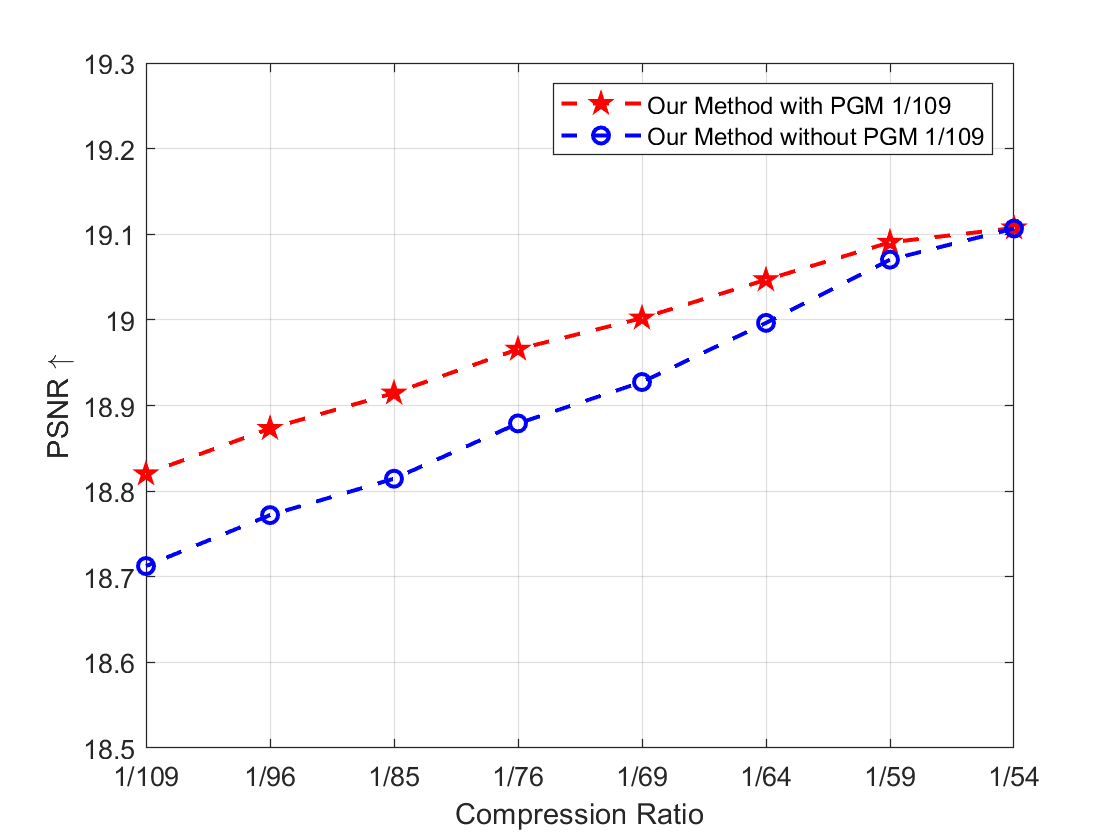}
        \caption{PSNR versus compression ratio, SNR=22dB.}
        \label{fig:7}
    \end{minipage}
\end{figure*}

\subsection{Channel Encoder and Decoder}
In this system, the compressed latent codes generated by the PGM cannot be transmitted as a whole. Due to the variability in images, the PGM may opt to ignore different features, and the number of neglected features may vary depending on the channel's state. To address this challenge, we employ 28 sets of channel encoders and decoders, each comprising a cascaded FC layer. All sets of encoders and decoders are encapsulated within a neural network, allowing for the invocation of the appropriate channel encoders and decoders for different numbers of dimensions of latent codes. We adopt an end-to-end training strategy to train the network and utilize Mean Squared Error (MSE) as the loss function. The loss function calculates the MSE between the compressed latent code and the reconstructed latent code.

\section{Simulation}
In this section, we evaluate the performance of proposed semantic communication system in terms of the transmission efficiency and robustness. We use CelebA-HQ dataset for training and testing. For simulating transmission on a real channel, we assume AWGN channel simulate the channels under different SNR. We use the existing semantic communication approach by \cite{ref6}, referred to as Deep JSCC, as the benchmark approach to compare to. We note that all approaches are trained and tested with the same datasets. During training, we set AWGN channel conditions with SNR randomly varying from -5dB to 5dB. 

The performance comparison of different approaches in terms of PSNR and LPIPS is presented in Fig.3 and Fig.4. A higher PSNR indicates better pixel-level restoration, while smaller LPIPS scores signify better objective perception restoration. It can be observed that with a 50\% compression ratio of Deep JSCC, our method achieves slightly lower PSNR scores but better LPIPS scores. This demonstrates that the proposed system preserves image perception well while endeavoring to maintain image details as much as possible. Furthermore, we find that PSNR and LPIPS values obtained by our approach remain almost unchanged under different SNRs, indicating the robustness of the proposed system to noisy channels.Fig.5 illustrates examples of reconstructed images produced by Deep JSCC and the proposed method, corroborating our results directly.

Additionally, we compare the performance of different compression ratios in terms of PSNR. During both training and testing, we set the Additive White Gaussian Noise (AWGN) channel conditions with an SNR of 5dB. The PSNR and LPIPS values obtained by our method, as depicted in Fig.6, also exhibit stability under varying compression ratios, further validating the robustness of our method to losses in transmitted information. Subsequently, we evaluate the effect of the PGM in Fig.7. During both training and testing, we maintain AWGN channel conditions with an SNR of 22dB. We observe that the PSNR of images decompressed by the PGM is significantly higher than those decompressed randomly, demonstrating that the PGM aids in reconstructing images with higher quality. Additionally, we note that the PSNR values in both situations are very close, indicating that the preserved features effectively control the style of the original images.

\section{CONCLUSION}
This work proposed a SemCom framework based on PGM, leveraging PGM as shared semantic knowledge between the transmitter and receiver for reducing transmission redundancy. We introduced a compression algorithm that discards negligible semantic information, considering the significance of semantic features and inference with PGM. The receiver then recovers the compressed semantic information through inference with the shared PGM. To evaluate the effectiveness of the proposed framework, we develop a PGM-based SemCom system for facial image transmission. Simulation results demonstrate the improvement in efficiency and robustness of the proposed semantic communication framework compared to existing methods.

\bibliographystyle{IEEEtran}
\bibliography{reference}

\end{document}